\documentclass[man,floatsintext]{apa7}

\usepackage{graphicx}
\usepackage{float}
\usepackage{minted}
\usepackage{adjustbox}
\usepackage[justification=centering]{caption}
\usepackage[square,numbers,sort&compress]{natbib}

\title{Predicting the Containment Time of California Wildfires Using Machine Learning}
\shorttitle{Predicting California Wildfires containment duration}
\author{Shashank Bhardwaj}
\authorsaffiliations{The University of Texas at Austin}
\date{October 2025}

\begin{document}

\maketitle
\section{Abstract}

California's wildfire season keeps getting worse over the years, overwhelming the emergency response teams. These fires cause massive destruction to both property and human life. Because of these reasons, there’s a growing need for accurate and practical predictions that can help assist with resources allocation for the Wildfire managers or the response teams. In this research, we built machine learning models to predicts the number of days it will require to fully contain a wildfire in California. Here, we addressed an important gap in the current literature. Most prior research has concentrated on wildfire risk or how fires spread, and the few that examine the duration typically predict it in broader categories rather than a continuous measure. This research treats the wildfire duration prediction as a regression task, which allows for more detailed and precise forecasts rather than just the broader categorical predictions used in prior work. We built the models by combining three publicly available datasets (a) California Department of forestry and fire protection's fire and resource assessment program (FRAP), (b) Data dictionary for FRAP, and (c) Dataset from California open data portal. This study compared the performance of baseline ensemble regressor, Random Forest and XGBoost, with a Long Short-Term Memory (LSTM) neural network. The ensemble models were trained using static features and aggregated weather data. We also trained the LSTM model on the same dataset with static features. The results show that the XGBoost model slightly outperforms the Random Forest model, likely due to its superior handling of static features in the dataset. The LSTM model, on the other hand, performed worse than the ensemble models because the dataset lacked temporal features. These results suggest that Tree-based ensemble models are best suited for situations where temporal features are unavailable and mostly static features are available. In contrast, LSTM models are more appropriate when temporal features such as meteorological data, wind speed and daily temperatures are available. Overall, this study shows that, depending on the feature availability, Wildfire managers or Fire management authorities can select the most appropriate model to accurately predict wildfire containment duration and allocate resources effectively.

\section{Introduction}

California's wildfire seasons are growing longer and more severe, posing a significant threat to lives, property, and natural resources \cite{CALFIRE2025FireSeason}, \cite{UCLA2025ClimateFireSeasons}, \cite{Stanford2020ExtremeFireWeather}. This also places a great strain on emergency teams. When a new fire is reported, emergency response managers face a series of critical decisions. One of the most fundamental question is how long it will take to contain the fire.

It is essential to make accurate and timely predictions of containment time for effective and strategic planning. These predictions help with crucial operational activities, including the allocation of firefighting resources (for example crews, engines, and aircraft), the planning of potential evacuations, and the management of public safety alerts. An underestimation of containment time can lead to a resource deficit, allowing a fire to grow uncontrollably. On the contrary, a significant overestimation can result in wasted or unnecessary diversion of critical resources from other impacted areas. 

California's fire problem has intensified dramatically in recent years. The state's five largest wildfires in recorded history have all occurred since 2018 \cite{Turco2023_AnthropogenicCC_ForestFires_CA}, \cite{calfire2025statistics}. Climate change has expanded the usual fire season into a year-round threat, while decades of fire suppression have left forests dense with combustible material \cite{Madakumbura2025_earlierFireSeason}. Recent mega-fires like the Camp Fire (2018) \cite{caloes2018camp} and the August Complex (2020) \cite{HernandezAyala2021_antecedentRainfallWildfireCA}, \cite{Campbell2022_NitrogenEmissionsWildfires}, \cite{PotterAlexander2022_SCUburnSeverity} burned for weeks, overwhelming suppression capabilities and forcing mass evacuations. These events underscore why accurate containment time predictions have become operationally essential.

Current practice relies heavily on expert judgment. Incident commanders estimate containment dates based on experience, visible fire behavior, and weather forecasts. While it is invaluable, this approach is very subjective and struggles to consistently process the complex, non-linear interactions of various environmental factors \cite{Paulus2022_DataCognitiveBiasCrisisInfo}. This is not just an informal process; it is often formalized in operational models. For instance, Fire-line production rates—a key variable in any containment forecast—are frequently estimated using "expert opinion" surveys of veteran crew bosses \cite{FriedGilless1989_firelineProduction}. Similarly, an incident commander's assessment of initial attack success is often supplemented by formal probability models built from the elicited judgment of fire management experts \cite{HirschCoreyMartell1998_fireCrewEffectiveness}. This reliance on experience is invaluable but also highlights an opportunity to complement human judgment with data-driven analytics. Machine learning offers a complementary tool, one that can process vast historical datasets to identify patterns human observers might miss.

In this research we built and evaluated a machine learning model capable of predicting the containment time for California wildfires. The specific objective was to forecast the total number of days between the initial fire report and the date of 100\% containment. By framing the problem as a continuous regression task, this study provided the granular, actionable predictions required for operational planning. This approach differs from existing work in three ways. First, we predict continuous duration in days rather than categorical bins (short/medium/long), providing the granular forecasts operations require. Second, we focus specifically on California's unique fire regime rather than generalizing across regions. Third, we test whether modeling temporal sequences of preignition weather improves predictions beyond static feature aggregation.

{\section{Background}}

Wildfire prediction is a complex problem. Over the years, a variety of approaches have been developed. Each of those have their own strengths and limitations. Traditional \textbf{physics-based models}, like FARSITE \cite{Zigner2020_FARSITE_SantaBarbara} and FlamMap \cite{Finney2006_FlamMapOverview}, predict how fires spread using environmental factors like fuel type, terrain, and weather conditions. These models can be very detailed and realistic. However, they often require extensive data and high computational resources, making them less practical for fast, large-scale predictions. For example, Peterson et al., 2011 attempted to reconstruct the wildfires in coastal regions of Santa Barbara County using wildfire models \cite{Peterson2011LongTermRegimes}
\begin{figure}[H]
\centering
\includegraphics[width=1\textwidth]{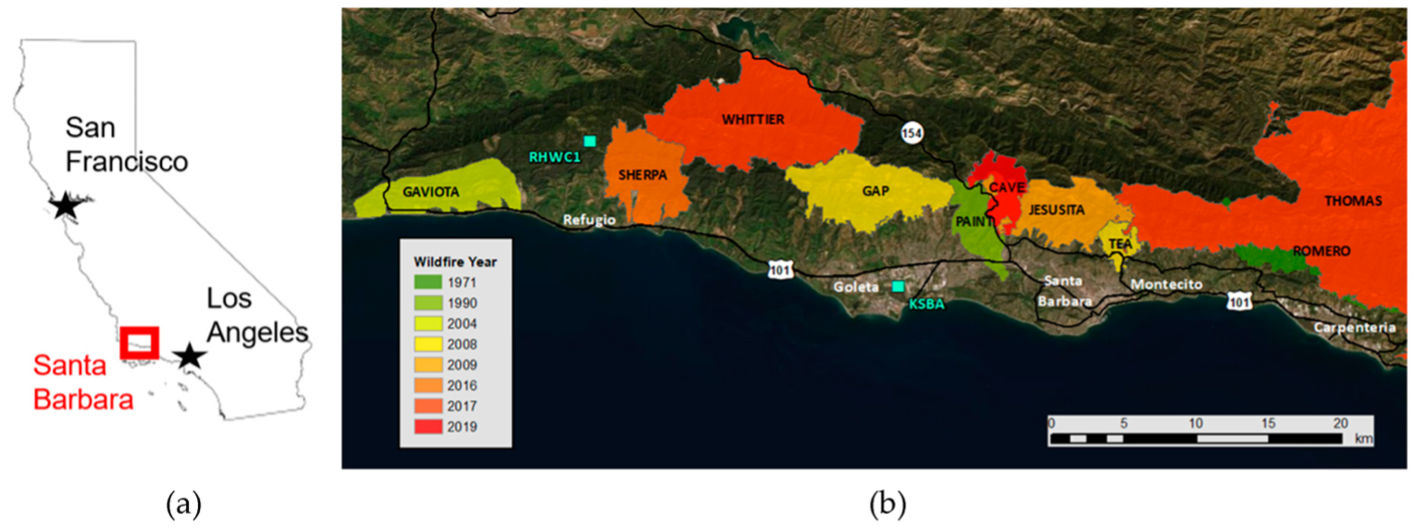}
\caption{Wildfire map of coastal region of Santa Barbara} \cite{Peterson2011LongTermRegimes}
\label{fig:wildfire-map-coastal-santa-barbara}
\end{figure}

In a parallel, statistical and probabilistic methods \cite{Preisler2004WildfireRisk} have been used to estimate outcomes such as burned area, fire risk, or containment likelihood. These approaches are usually faster to run and easier to interpret. But their simplicity comes at a cost: they often struggle to capture the complex, non-linear relationships that govern how a fire behaves under changing weather, variable terrain, or differing vegetation types.
\begin{figure}[H]
    \centering
    \includegraphics[width=1\linewidth]{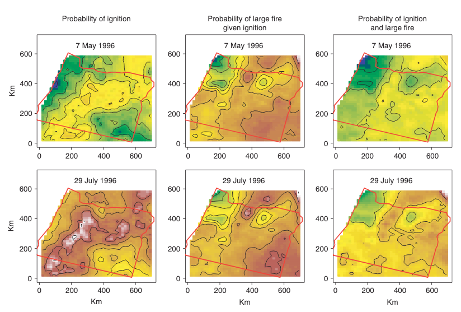}   
    \caption{Wildfire pattern over different terrain} \cite{Preisler2004WildfireRisk}
\end{figure}
Even with these approaches, containment time varies widely with fire size. The following plot which is generated from the California historical fire parameters records, illustrates the relationship between historical California wildfire sizes and their containment times \cite{CALFIRE_HistoricalPerimeters}. Each point represents a single fire, with the x-axis showing log-transformed fire size (acres) and the y-axis showing the number of days until full containment. The plot highlights the wide variability in containment durations - even among fires of similar sizes - underscoring the challenges of forecasting containment using simple rules or expert judgment alone
\begin{figure}[H]
    \centering
    \includegraphics[width=1\linewidth]{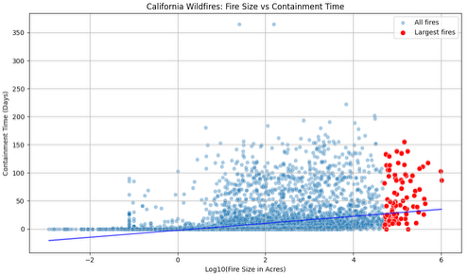}
    \caption{Log transformed fire size (in acres) with Containment duration}
\end{figure}

More recently, machine learning (ML) methods have become popular in wildfire research. ML models are particularly useful because they can learn patterns directly from large historical datasets, without requiring explicit modeling of every physical process. For example, ML has been used to predict fire risk and ignition \cite{Pham2022_CAWildfirePrediction}, forecast fire spread \cite{Marjani2023FirePred}, or classify fire duration \cite{Tzimas2024ForestFireDuration}.

But, even with these advances, most studies focus on risk assessment or broad categories rather than predicting continuous containment time. In California, where fires can last for days or even weeks, this kind of fine-grained prediction is especially valuable. 

There is also increasing interest in deep learning techniques. Recurrent neural networks, such as LSTM and GRU, are well-suited to modeling temporal patterns, like daily weather changes affecting fire growth. Convolution neural networks can analyze spatial patterns in satellite images to detect hot-spots or burned areas. Some researchers are even experimenting with graph-based models to capture interactions between neighboring regions \cite{das2025graph}, \cite{zhao2024causal}, \cite{michail2025firecastnet}. Despite their promise, these methods have rarely been applied to predicting exact containment duration, leaving a clear gap that this research seeks to address. \newline

\subsection{Related Works \& Literature Review}

Previous research can be broadly categorized into three main areas. First, several studies have focused on Predicting Wildfire Risk and Ignition \cite{Pham2022CaliforniaWildfireML}. These models analyze historical and environmental factors - such as vegetation health, land surface temperature, and historical fire patterns to generate spatial risk maps. Classification models, including Random Forest and Support Vector Machines (SVM), have demonstrated strong performance in identifying areas as high-risk or low risk for a new ignition \cite{illarionova2025exploration}, \cite{Chen2024ForestFireMapping}, \cite{Ali2025ForestFireRF}. While valuable for long-term planning, these models do not address the dynamics of a fire after it has started.

Another research has focused on Predicting Wildfire Spread \cite{Marjani2023FirePred}. These models are concerned with forecasting the growth and spread dynamics of a fire after ignition, often predicting the total burned area over a specific time horizon. Research in this domain has found that while traditional statistical methods can provide baseline predictions, ensemble methods like XGBoost offer superior performance. This is particularly true when incorporating complex, time-varying inputs, such as meteorological data.

The work most closely related to this proposal involves Predicting Wildfire Duration as a Category. A recent study on forest fires in Greece employed a Random Forest classifier to predict fire duration \cite{Tzimas2024ForestFireDuration}. This model achieved high accuracy (87-92\%) but framed the problem as a multi-class classification task, binning fires into three duration categories: short, medium, and long. Similar studies have been done outside Greece as well \cite{Pérez-Porras2021LargeWildfires} \cite{Angelov2021WildfireSizeBinary} \cite{caron2025localized} \newline

\subsection{Research Gap}

While these existing studies provide a valuable foundation, a significant gap remains. To our knowledge, no prior work has specifically focused on predicting the “continuous containment time” (measured in days) as a regression problem, particularly for California wildfires. This research project addresses this gap in three unique ways. First, it adopts a geographic focus\textbf{: }while the most comparable duration study focused on Greek forest fires, California's distinct Mediterranean climate, diverse and complex terrain, and unique fire regimes necessitate region-specific modeling to capture local fire behavior patterns. Second, it proposes a regression-based problem formulation, predicting continuous days rather than classification or categorical bins. This approach is more challenging but provides the granular, quantitative predictions required for operational resource planning and logistics. Third, it emphasizes on data integration\textbf{,} combining incident records, meteorological data, and topographic/fuel data to capture the multifaceted drivers of fire containment.\newline

\section{DATA \& VARIABLES}

For this study, we gathered data from multiple sources and joined them to create a master dataset. After that, we cleaned the data, removed rows with empty or missing columns. The dataset retrieved from official historical California wildfire website (Fire and Resource Assessment Program) \cite{CALFIRE_PerimetersAll} contains wildfire events from around 1878 to January 2025. Latitude and Longitude values of the fire event location is captured from the Shape files \cite{CaliforniaOpenData_Shapefile} and joined with the master dataset. The data is joined using common columns “Year\_“ and “IRWIN ID”. The dataset also contains “Cause” column containing categorical values. The mappings are downloaded from the data dictionary.

\subsection{Data Sources}
The dataset used in this study is a composite of three data sources:

\subsubsection{California Department of Forestry and Fire Protection's Fire and Resource Assessment Program (FRAP)}
This dataset contains the historical fire data (CAL FIRE California Fire Perimeters - all) \cite{CALFIRE_PerimetersAll} . This dataset provides the alarm and containment dates along with the cause (categorical numerical values) and other essential details such as GIS\_ACRES, Fire name, Direct protection agency etc. This data is combined with other datasets containing the investigated cause of ignition (actual value), geospatial information such as Latitude \& Longitude. The combined dataset helped derive the target variable Containment\_Time\_Days.

\subsubsection{Data Dictionary for FRAP}
This contains the actual values for cause of ignition, Direct protection agency responsible for fire, method used to collect perimeter data (‘C\_Method’ column in dataset) \cite{FRAP_DataDictionary}

\subsubsection{California Open Data Portal}
This contains Shapefile having Latitude and Longitude of the fire events \cite{CaliforniaOpenData_Shapefile}. This also contains the columns such as ‘IRWIN ID’, ‘Fire Name’ and Year which are used to join the data with previous datasets to generate a master dataset.\newline

\subsection{Data Preparations and Descriptive Statistics }

We downloaded the initial dataset used in this study from the California Department of Forestry and Fire Protection's Fire and Resource Assessment Program (FRAP). This dataset contains the historical wildfire perimeter records across California state. The raw data contain several inconsistencies and historical artifacts that require pre-processing before it can be used for predictive modeling.

\subsubsection{Data Cleaning and Integrity checks }
The FRAP dataset includes date fields, agency identifiers, incident descriptors and polygon geometric for each recorded fire incident. The first stage of preparation involved standardizing field types and identifying the unusable records such as missing or corrupted data.

\paragraph{Date Standardization}
Both ALARM\_DATE and CONT\_DATE fields were parsed into date-time object. Rows having non-parseable dates were removed.

\paragraph{Inconsistent fields}
Records with inconsistent values such as CONT\_DATE lower than ALARM\_DATE were removed.

\paragraph{Removal of duplicate and malformed records}
Some entries contained duplicate IRWINIDs or incomplete polygon geometries. Duplicate rows and entries with missing geometries were removed. This ensured that each incident record represented a unique fire incident with complete spatial and temporal information.

\subsubsection{Spatial Preprocessing}
The shapefile provided by California Open Data portal includes detailed polygons for the burned area for each incident. To create the usable spatial features, centroid coordinates (latitude and longitude) were extracted from each polygon. These coordinates were merged with the cleaned tabular data using the unique incident identifier (IRWIN ID and FIRE NAME). Entries for which centroid calculation failed due to corrupted records were excluded. 

\subsubsection{Descriptive Statistics }
After completing data cleaning, removal of invalid entries, and merging the FRAP perimeter attributes with centroid coordinates extracted from the shape-files, the final analytical dataset consisted of \textbf{15547 wildfire incidents} with complete spatial, temporal, and descriptive information.

The dataset spans multiple decades of California wildfire activity. Because the goal of this study is to evaluate model performance on future events, a temporal split was used: fires occurring before 2018 were used for training, and fires from 2018 onward were reserved as validation or test set. This ensures that evaluation reflects generalization to unseen future fires rather than random resampling.

Table 1 reports descriptive statistics for the main numerical variables used in modeling. The strong difference between the mean and median containment duration confirms the heavy right-skew of the target variable and supports the use of a log-transformed target during model training 

\begin{table}[H]
\centering

\begin{tabular}{|l|l  |l  |l  |l  |l  |l  |l  |l |}
\hline
Variable & count & mean  & std  & min  & 25\% & 50\% & 75\% & max \\
\hline
containment\_days & 15546 & 5.787 & 21.128 & 0.00000 & 0.00000 & 0.00000 & 2.000000 & 638 \\
\hline
log\_cont\_days& 15546 & 0.699 & 1.1648 & 0.00000 & 0.00000 & 0.00000 & 1.098612 & 6.45 \\
\hline
GIS\_ACRES & 15546 & 2138.36 & 17391.38 & 0.001171 & 21.91567 & 106.392 & 505.1853 & 1032700 \\
\hline
log\_acres & 15546 & 4.77410 & 2.2668 & 0.001170 & 3.131821 & 4.676486 & 6.226903 & 13.85 \\
\hline

\end{tabular}

\end{table}

\subsection{Target Variable Derivation and Pre-Processing}

The primary target for this regression task is to calculate the Containment Time (in days). This variable is computed as the number of days between a fire's Ignition Date and its 100\% Containment Date. The combined FRAP perimeter dataset contains two date fields which are important for this calculation: “Alarm\_Date” which represents the recorded fire ignition date and “Cont\_Date” which represents the recorded date when the fire was fully contained. Both fields stored as strings in the dataset were converted to date-time objects for the calculation. Records or rows where these dates were either missing or corrupted were removed to avoid any parsing errors. Also, the records where there were structural inconsistencies such as Cont\_Date value lower than Alarm\_date were removed to avoid distorting the model behavior.

One important nuance of the combined FRAP dataset is that the “YEAR\_” column does not always reflect the true year of the wildfire. This represents the year when the fire incident was digitized. So, for the very old wildfire events (e.g. year 1899), the YEAR\_ value shows a later date (date when digitized) e.g. 1945 but the Alarm and Containment dates shows thew actual date i.e. 1899.

The resulting variable, i.e. containment\_days is a continuous measure representing the operational duration of suppression activity for each fire. This metric forms the basis for all predictive modeling tasks in this study. 

Analysis of this variable shows a strong positive (right) skew. This is typical for duration-related wildfire metrics, where most incidents are resolved quickly while a smaller number persists for extended periods. This skewness is also consistent with the broader pattern observed in wildfire-related variables – such as burned area – which are known to follow heavy-tailed distributions \cite{Li2021SpatialTemporalCA}.

\begin{figure}[H]
    \centering
    \includegraphics[width=1\linewidth]{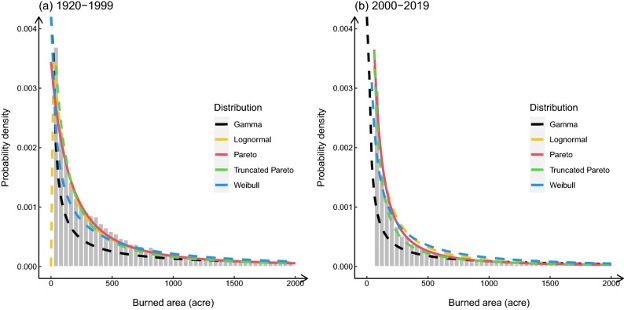}
    \caption{Wildfire data showing heavy-tailed distribution} \cite{Li2021SpatialTemporalCA}
\end{figure}

The dataset contains many small, quickly contained fires (1-3 days) and a long tail of large, complex "megafires" that burn for many weeks. Training a regression model on such a skewed distribution can lead to high sensitivity to outliers and poor overall performance, as the model would be heavily biased by the few extreme-duration events.

To address this and stabilize the variance, a log-transformation was applied. The final target variable used for all model training is log\_Containment\_Time, calculated as 
$$
target = log(1 + Containment\_Time\_Days\})
$$
All model predictions are generated in this log-transformed space and are then converted back to the original scale for evaluation and interpretation in actual days using the inverse-transformation.
$$
prediction = exp(target) - 1
$$
This ensures that reported errors and model comparisons reflect actual containment durations in days.

Plotting the log transformed image looks much more interpretable as opposed to regular plot.

\begin{figure}
    \centering
    \includegraphics[width=1\linewidth]{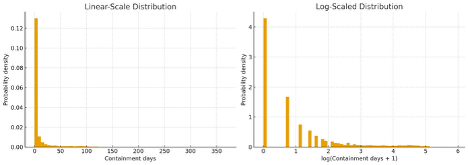}
    \caption{Linear-Scale vs Log-Scale distribution of wildfire containment days}
\end{figure}

\subsection{Feature Engineering }

For the feature engineering in this study, we constructed a set of meaningful and interpretable variable derived directly from the FRAP dataset enriched with coordinates from the corresponding shape-files. Because the dataset provides only static attributes for each incident, all engineered features represent fire-level characteristics. These features capture seasonal timing, spatial location, fire size and incident-level description attributes. All of these have established relationships with wildfire duration and suppression outcomes.

\subsubsection{Temporal Features}

Although the YEAR\_ and DECADES fields are present in the raw data, they were excluded due to inconsistencies arising due to historical digitization practices. Both YEAR\_ and DECADES columns represent the year or decade when the records were digitized rather than the actual year or decade of wildfire. Instead, all temporal features were derived directly from the ignition date (ALARM\_DATE), which provides the most reliable order of the incident.

\subsubsection{Spatial Features}

Latitude and Longitude fields were generated by computing the centroids of the fire perimeters from the shape-file. These coordinates provide useful spatial representation of each incident’s location. Spatial attributes play an important role in wildfire behavior because topography, fuel type and climate vary significantly across California. \cite{Syphard2024_geography_fire_patterns} \cite{AlBashiti2022_machineLearningWildfire}

\subsubsection{Fire Characteristics}

The FRAP dataset includes various static descriptors of fire behavior and management: (1) burned\_area\_acres (GIS\_ACRES) – total area burned, (2) log\_acres = log (1 + GIS\_ACRES) – Since burned area is also highly skewed, a log-transformed version is more useful to reduce the influence of large incidents and help stabilize the variance. \cite{Wang2020_wildfire_burned_area} \cite{Coffield2019_fire_size_prediction} \cite{Li2023_AttentionFire}

\subsubsection{Categorical Descriptive Features}

FRAP dataset also has Various categorical attributes which capture causal aspects of each fire incident that may influence the containment time: (1) CAUSE – ignition cause classification \cite{Balch2017_humanStartedWildfires},  \cite{Fusco2022_detectionBiasesWildfire}, (2) AGENCY - Direct protection agency responsible for fire, (3) UNIT\_ID - ICS code for unit, (4) C\_METHOD - Method used to collect perimeter data, (5) OBJECTIVE - Tactic for fire response

\subsubsection{LSTM-Specific Considerations}

Although LSTM models are traditionally applied to sequential data, the architecture can also be used as a deep nonlinear mapping when fed with fixed-length vectors. For this reason, the same static feature set described above was provided to the LSTM in vectorized form, with categorical variables embedded and numeric variables standardized. No temporal sequences were available in the dataset, so the LSTM effectively learns higher-order interactions among static features rather than time-series patterns. The inclusion of the LSTM model thus serves as a comparative deep learning baseline alongside the tree-based methods.

\section{METHODS}

We used the FRAP dataset for modeling the California wildfire containment duration. The modeling pipeline consists of: (1) dataset preparation (clean and transform), (2) pre-processing \& encoding, (3) training the models, (4) evaluating the models on the validation dataset, (5) Finally, comparing the performance of those models. For the purpose of this study, we selected Random Forest and XGBoost as the baseline ensemble models. Also used a deep learning LSTM model.

\subsection{Target variable}

Containment duration was calculated directly from ALARM\_DATE and CONT\_DATE and used as the prediction target. Due to its heavy right-skewed distribution, a log transformation was applied during model training, with predictions converted back to days using:
$$
y\_days = exp(log\_containment\_days) - 1
$$
This transformation ensured numerical stability across all modeling approaches.

\subsection{Training Strategy}

To train the model, we cleaned and preprocessed FRAP dataset, then split into training and test set. This split was chosen according to temporal split strategy rather than random sampling. Wildfire data for incidents happened before the year 2018 was used for training and between 2018 to 2025 was used for validation set. This also helped reflect the real scenario where model predicts future events rather than identifying the past events. We used scikit-learn library in Python to build the Tree-based ensemble models, while the LSTM was built using the Keras (TensorFlow) library. Year 2018 was taken as a threshold for train/test split for all model evaluations. Hyper-parameters were tuned using 5-fold cross-validation. A separate validation set is not required here since we used GridSearchCV for training which uses train data as folds and rotate the validation set internally to tune the hyper-parameters. Total training samples came out to be 12804 (records until the year 2018) and test samples as 2742 (records between the year 2018 and 2025).\newline

\subsection{Random Forest Regression - Baseline}

We selected Random Forest as an initial benchmark due to its ability to model mixed feature types and interpretability via feature importance. Also, because Tzimas et al. (2024) achieved strong results using for fire duration classification with this model, suggesting the algorithm can capture relevant patterns \cite{Tzimas2024ForestFireDuration}. Hyper-parameters tuned during training included: (a) number of trees (n\_estimators), (b) maximum tree depth, (c) minimum samples per split/leaf, (d) bootstrap sampling

We used RandomForestRegressor from Scikit-learn library here for training the Random Forest model. Random Forest served as a stable baseline against which more advanced models were compared. Following hyperparameter were tuned: n\_estimators [200, 350, 500, 700],  max\_depth [10, 20, 40, None],  min\_samples\_split [2, 5, 10],  min\_samples\_leaf [1, 2, 4],  max\_features ["sqrt", "log2", None]

After training with the current FRAP dataset, the optimal values of hyper-parameters came out to be: n\_estimators 350,  max\_depth 10,  min\_samples\_split 2,  min\_samples\_leaf 1,  max\_features "sqrt". The overall mean absolute error for Random Forest model came out to be 6.63 and Root mean square error as 22.60. This was slightly higher than XGBoost which means this model underperformed. \newline

\subsection{XGBoost Regression}

XGBoost (Extreme Gradient Boosting) is selected as another model because it works better on the structured tabular dataset. Secondly, XGBoost can model non-linear interactions between spatial, temporal and categorical wildfire characteristics. Marjani et al. (2023) found XGBoost superior to other methods for California wildfire spread prediction \cite{Marjani2023FirePred}. Other studies also found it to be better \cite{chen2016xgboost}. Gradient boosting builds decision trees in multiple iterations correcting the residual of prior iteration. This enables higher accuracy maintaining the generalizations. For this study, we tuned following hyper-parameters: (a) learning rate, (b) max\_depth, (c) number of trees, (d) subsample and column sampling rates, (e) early stopping rounds.

For the XGBoost model, we used the standalone library providing the model as XGBRegressor. Following hyper-parameters were tuned: n\_estimators: [300, 500, 700], learning\_rate: [0.01, 0.05, 0.1], max\_depth: [4, 6, 10], subsample: [0.6, 0.8, 1.0], colsample\_bytree: [0.6, 0.8, 1.0], gamma: [0, 0.2, 0.4]

After training, following hyper-parameter values came out to be the most optimal:  n\_estimators: 500, learning\_rate: 0.01, max\_depth: 4, subsample: 0.6, colsample\_bytree: 0.8, gamma: 0.4. The overall mean absolute error for XGBoost model was 6.53 and Root mean square error as 22.37. This model performed slightly better than Random Forest.

Due to the nature of wildfire attributes, with heavy-tailed response (containment duration), XGBoost was expected to outperform the Random Forest model. Although, it was only marginally better. \newline

\subsection{LSTM Deep learning Model}

Long Short-Term Memory (LSTM) networks are design for sequential or evolving signals such as fire spread over time \cite{hochreiter1997long}. The dataset we chose did not have temporal sequences such as wind speed or other weather conditions prior to fire alarm date. We ran 10 trials of BayesianOptimization tuning the following hyper-parameters: units, dropouts and learning rate. After completing all the trials, the most optimal values for hyper-parameters came out to be:  units = 192, dropouts = 0.1 and learning rate = 0.001. The overall mean absolute error for LSTM model was 7.07 and Root mean square error as 23.67. 

Due to the limited features in the dataset, LSTM did not perform better than XGBoost and exhibited the weakest performance. But for future work, we can incorporate daily meteorological data, fuel moisture history, vegetation time series, or Satellite heat signatures along with the current static dataset to train the LSTM model which should outperform the tree-based models. \newline

\subsection{Evaluation Metrics}

Model performance was assessed using three metrics computed on real units (inverse transformed i.e. actual number of days instead of logarithmic data) predictions: (a) Mean absolute error (MAE), (b) Root mean squared error (RMSE), (c) R-squared or proportion of variance. \newline

\section{Results \& Analysis}

All the three models (Random Forest, XGBoost and LSTM) were tested on the test set to compare the performance when compared with the actual values. The comparison was performed on the real units (number of days) after converting the model output back from the log values. 

\subsection{Overall Model performance}

Following table summarizes the performance of the trained models. XGBoost achieved the strongest predictive accuracy among the three models and across all the metrics. Random forest performed better than LSTM mostly due to the absence of temporal parameters.
\begin{table}
    \centering
    \begin{tabular}{|c|c|c|c|}\hline
         Model&  MAE&  RMSE& R²\\\hline
         Random Forest&  6.6317&  22.6070& 0.1897\\\hline
         XGBoost&  6.5261&  22.3712& 0.2065\\\hline
         LSTM&  7.0661&  23.6704& 0.1117\\ \hline
    \end{tabular}
    
\end{table}

XGBoost achieved the lowest MAE and RMSE. This indicates that XGBoost was consistently better on both regular as well as high-duration fires. The R-square value further suggests stronger performance. This aligns with the literature studies of structured wildfires-related data where the parameters are static, and time related dimensions are unavailable.

\subsection{Error distribution \& Model behavior}

Residual analysis of all three models demonstrates a consistent pattern: (a) Short duration fires (0-10 days) were predicted with reasonable accurately, which also forms the majority of the dataset, (b) Errors increased progressively with containment duration, specifically for incidents extending over \~40 - 50 days, (c) Extreme long tail events remains difficult to predict

A visualization of prediction error versus true containment duration shows that all models systematically underestimate long-duration fires with prediction error increasing monotonically with containment time. This highlights the difficulty in predicting heavy-tail extreme wildfires using only initial-condition attributes.

\begin{figure}
    \centering
    \includegraphics[width=1\linewidth]{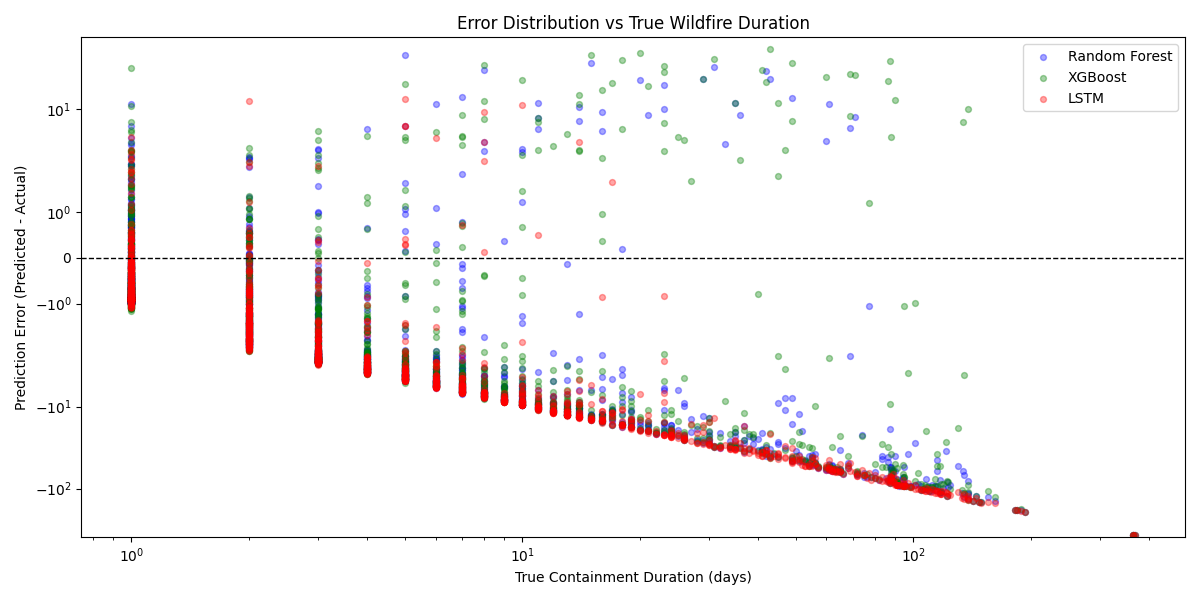}
    \caption{Prediction error for containment duration for various models}
\end{figure}

Random Forest and XGBoost both exhibit extreme positive prediction errors, indicating some overestimation but most of it was underestimation. LSTM on the other hand displays the least variance in it's prediction. It produced compressed outputs near the mean which shows it's stability but poor adaptability to extreme duration estimation.

This is an expected outcome when predicting a variable with fat-tailed behavior using static incident attributes alone.

\subsection{Why XGBoost Outperformed LSTM}

The current dataset provides only static parameters per wildfire incident. Without multi-day fire spread, fuel moisture evolution, or temporal meteorological context, an LSTM behave like a dense feedforward network. XGBoost on the other hand naturally handles non-linear tabular interactions, and heavy-tailed regression targets like containment days in current dataset. This validates the hypothesis: In static fire-incident data, gradient-boosted trees outperform recurrent neural networks. The LSTM results are still valuable. It demonstrates that deep learning does not under-perform due to model weakness but rather due to missing temporal features in the dataset.

\subsection{Feature importance \& Interpretation (Tree based models)}

Feature importance analysis was performed using both Random Forest and XGBoost models. Figure 7 and 8 shows the graphs plotted for all the features in these two models. Feature importance ranking for these models shows which factors most influenced containment duration

\begin{figure}
    \centering
    \includegraphics[width=0.75\linewidth]{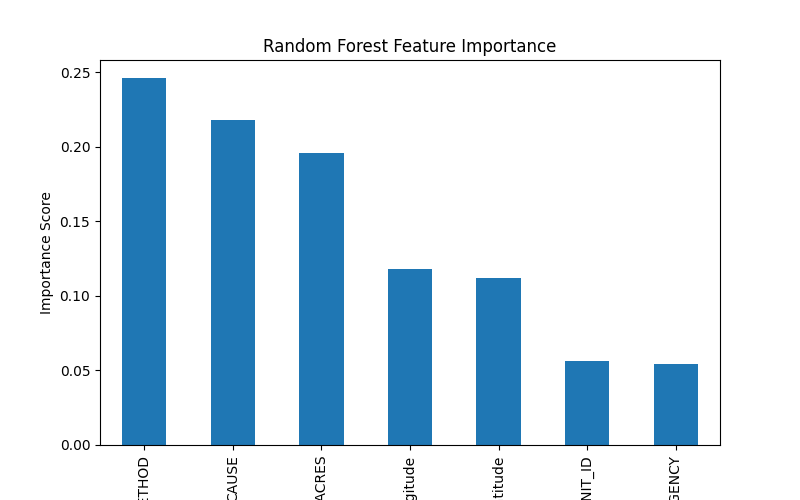}
    \caption{Feature Importance for Random Forest Model}
\end{figure}

\begin{figure}
    \centering
    \includegraphics[width=0.75\linewidth]{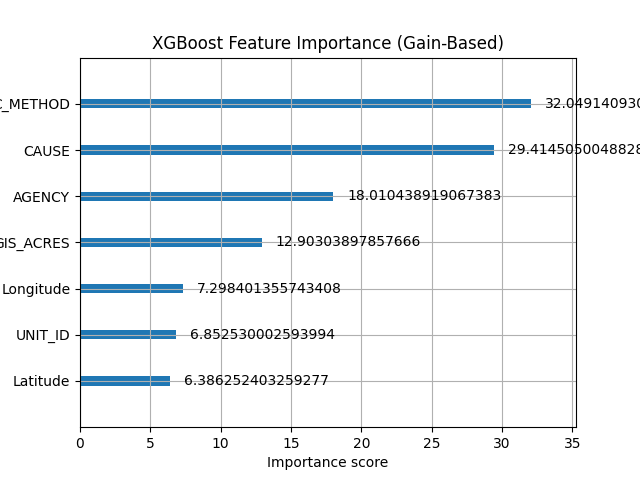}
    \caption{Feature Importance for XGBoost model}
\end{figure}

Both models (RF and XGB) independently ranked C\_Method (Perimeter collection method) as the strongest predictor of containment time. It's predictive strength likely arise because advanced mapping technologies are more commonly deployed for larger and harder to contain incidents which span multi-weeks. Therefore, C\_Method acts as a proxy variable for event scale and operational complexity. CAUSE and GIS\_ACRES are next strong predictors which are both naturally tied to wildfire behavior. Geographic coordinates like Latitude and Longitude hold moderate explanatory power. Remaining parameters like UNIT\_ID contributed weakly which indicates that the containment duration is driven more by physical \& situational characteristics rather than formal authority boundaries.

\subsection{Practical Implications}

These results indicate that short-duration (1-10 days) fire incidents which forms the majority of the fire incidents can be predicted with reasonable accuracy using static attributes. However, the long duration extreme cases (40 - 50 days) remain difficult to predict due to fat-tailed nature of containment time. Gradient boosted tree models (e.g. XGBoost) are best suited for static non temporal wildfire attributes and outperform deep learning models under this constraint. Deep learning models performance is limited without sequential progression data (weather evolution, suppression actions over time, fuel moisture change, etc.), but is expected to improve significantly when temporal features are incorporated.

For practical considerations, what this means is that tree-based gradient-boosted models provide useful predictions for early incident planning and resource allocation for a fire incident. Whereas the deep learning architectures can assist with multi-day containment forecasting once the temporal data attributes are made available.

\subsection{Limitations and future directions}

The key limitation here is the lack of temporal features. If we can join the current dataset with temporal parameters such as (a) daily progression perimeter growth, (b) meteorological sequences like wind speed, temperature etc., (c) Fuel moisture indices, (d) Satellite thermal evolution, then LSTM or GRU models could capture the evolving fire behavior more effectively than XGBoost.

\section{Summary}

The current study addressed a critical gap in wildfire management research by building a machine learning model to predict the California wildfire containment duration in days. XGBoost provided the best performance for predicting wildfire containment duration from static FRAP attributes. Random Forest served as a stable baseline, while LSTM performance confirmed that recurrent deep models require temporal structure to fully realize their strengths. These findings established a benchmark for a practical prediction method for California wildfire containment duration. It also presents a clear path forward for future data collection and modeling work.

\bibliographystyle{IEEEtran}
\bibliography{references}   

@misc{CALFIRE2025FireSeason,
  author       = {{CAL FIRE}},
  year         = {2025},
  title        = {2025 Fire Season Incident Archive},
  howpublished = {Official Report},
  note         = {Wildfire seasons trending above normal with increased threat to structures and resources},
  url          = {https://www.fire.ca.gov/incidents/2025},
}

@article{UCLA2025ClimateFireSeasons,
  author  = {{UCLA Newsroom}},
  year    = {2025},
  title   = {Human-caused climate change is expanding California's fire seasons},
  journal = {UCLA Newsroom},
  note    = {Fire seasons advanced 6--46 days earlier (1992--2020)},
  url     = {https://newsroom.ucla.edu/releases/human-caused-climate-change-expanding-california-fire-seasons},
}

@misc{Stanford2020ExtremeFireWeather,
  author       = {{Stanford Sustainability}},
  year         = {2020},
  title        = {Longer, more extreme wildfire seasons},
  howpublished = {Research Summary},
  note         = {Autumn extreme fire weather days doubled since 1980s},
  url          = {https://sustainability.stanford.edu/news/longer-more-extreme-wildfire-seasons},
}

@article{Madakumbura2025_earlierFireSeason,
  title        = {Anthropogenic warming drives earlier wildfire season onset in California},
  author       = {Madakumbura, Gavin D. and Moritz, Michael A. and McKinnon, Karen A. and Williams, A. Park and Rahimi, Seyed and Bass, Brian and Norris, Jeffrey and Fu, Rong and Hall, Alex},
  journal      = {Science Advances},
  volume       = {11},
  number       = {32},
  pages        = {eadt2041},
  year         = {2025},
  doi          = {10.1126/sciadv.adt2041},
  url          = {https://www.science.org/doi/10.1126/sciadv.adt2041}
}

@misc{calfire2025statistics,
  title={Statistics},
  author={{CAL FIRE}},
  year={2025},
  url={https://www.fire.ca.gov/our-impact/statistics},
  organization={California Department of Forestry and Fire Protection}
}

@article{Turco2023_AnthropogenicCC_ForestFires_CA,
  title        = {Anthropogenic climate change impacts exacerbate summer forest fires in California},
  author       = {Turco, Marco and Abatzoglou, John T. and Herrera, Sixto and Zhuang, Yizhou and Jerez, Sonia and Lucas, Donald D. and AghaKouchak, Amir and Cvijanovic, Ivana},
  journal      = {Proceedings of the National Academy of Sciences of the United States of America},
  volume       = {120},
  number       = {25},
  pages        = {e2213815120},
  year         = {2023},
  doi          = {10.1073/pnas.2213815120},
  url          = {https://www.pnas.org/doi/10.1073/pnas.2213815120}
}

@techreport{caloes2018camp,
  title={2018 Camp Fire After Action Report},
  author={{California Governor's Office of Emergency Services (Cal OES)}},
  year={2025},
  institution={Cal OES},
  url={https://www.caloes.ca.gov/wp-content/uploads/Preparedness/Documents/FINAL-AAR-2018-Camp-Fire-508-Clean-Copy-11.17.25.pdf}
}

@article{HernandezAyala2021_antecedentRainfallWildfireCA,
  title        = {Antecedent Rainfall, Excessive Vegetation Growth and Its Relation to Wildfire Burned Areas in California},
  author       = {Hern{\'a}ndez Ayala, J. J. and Mann, J. and Grosvenor, M.},
  journal      = {Earth and Space Science},
  volume       = {8},
  number       = {9},
  pages        = {e2020EA001624},
  year         = {2021},
  doi          = {10.1029/2020EA001624},
  url          = {https://agupubs.onlinelibrary.wiley.com/doi/full/10.1029/2020EA001624}
}

@article{Campbell2022_NitrogenEmissionsWildfires,
  title        = {Pronounced increases in nitrogen emissions and deposition due to the historic 2020 wildfires in the western U.S.},
  author       = {Campbell, Patrick C. and Tong, Daniel and Saylor, Rick and Li, Yunyao and Ma, Siqi and Zhang, Xiaoyang and Kondragunta, Shobha and Li, Fangjun},
  journal      = {Science of the Total Environment},
  volume       = {839},
  pages        = {156130},
  year         = {2022},
  doi          = {10.1016/j.scitotenv.2022.156130},
  url          = {https://doi.org/10.1016/j.scitotenv.2022.156130}
}

@article{PotterAlexander2022_SCUburnSeverity,
  title        = {Machine learning to understand patterns of burn severity from the SCU Lightning Complex Fires of August 2020},
  author       = {Potter, Christopher and Alexander, Olivia},
  journal      = {California Fish and Wildlife Journal},
  volume       = {108},
  number       = {1},
  pages        = {108--120},
  year         = {2022},
  doi          = {10.51492/cfwj.108.6},
  url          = {https://journal.wildlife.ca.gov/2022/05/13/machine-learning-to-understand-patterns-of-burn-severity-from-the-scu-lightning-complex-fires-of-august-2020/}
}

@mastersthesis{das2025graph,
  title = {A Graph Neural Network Based Approach for Predicting Wildfire Burned Area},
  author = {Das, Ursula},
  school = {University of Waterloo},
  type = {MASc thesis},
  year = {2025},
  note = {Electrical and Computer Engineering},
  advisor = {Naik, Kshirasagar},
  url = {https://uwspace.uwaterloo.ca/items/70bea407-594e-4fa7-af82-178d9b8d04d4},
  urldate = {2025-11-30},
  handle = {hdl:10012/21459}
}

@article{zhao2024causal,
  author = {Zhao, Shan and Prapas, Ioannis and Karasante, Ilektra and Xiong, Zhitong and Papoutsis, Ioannis and Camps-Valls, Gustau and Zhu, Xiao Xiang},
  title = {Causal Graph Neural Networks for Wildfire Danger Prediction},
  journal = {arXiv preprint arXiv:2403.08414},
  year = {2024},
  month = {mar},
  note = {Accepted by ICLR 2024 Machine Learning for Remote Sensing (ML4RS) Workshop},
  doi = {10.48550/arXiv.2403.08414},
  url = {https://arxiv.org/abs/2403.08414},
  urldate = {2025-11-30}
}

@article{michail2025firecastnet,
  author = {Michail, Dimitrios and Davalas, Charalampos and Panagiotou, Lefki-Ioanna and Prapas, Ioannis and Kondylatos, Spyros and Bountos, Nikolaos Ioannis and Papoutsis, Ioannis},
  title = {FireCastNet: Earth-as-a-Graph for Seasonal Fire Prediction},
  journal = {arXiv preprint arXiv:2502.01550},
  year = {2025},
  month = {feb},
  doi = {10.48550/arXiv.2502.01550},
  url = {https://arxiv.org/abs/2502.01550}
}

@article{illarionova2025exploration,
  author  = {Illarionova, Svetlana and Shadrin, Dmitrii and Gubanov, Fedor and Shutov, Mikhail and Tasuev, Usman and Evteeva, Ksenia and Mironenko, Maksim and Burnaev, Evgeny},
  title   = {Exploration of geo-spatial data and machine learning algorithms for robust wildfire occurrence prediction},
  journal = {Scientific Reports},
  year    = {2025},
  volume  = {15},
  pages   = {94002},
  doi     = {10.1038/s41598-025-94002-4},
  url     = {https://www.nature.com/articles/s41598-025-94002-4}
}

@article{Chen2024ForestFireMapping,
  author    = {Xinbao Chen and Yaohui Zhang and Shan Wang and Zecheng Zhao and Chang Liu and Junjun Wen},
  title     = {Comparative study of machine learning methods for mapping forest fire areas using Sentinel-1B and 2A imagery},
  journal   = {Frontiers in Remote Sensing},
  volume    = {5},
  pages     = {1446641},
  year      = {2024},
  doi       = {10.3389/frsen.2024.1446641},
  url       = {https://www.frontiersin.org/journals/remote-sensing/articles/10.3389/frsen.2024.1446641/full}
}

@article{Ali2025ForestFireRF,
  author    = {Shahwan younis ali and Omar Sedqi Kareem},
  title     = {Forest Fire Prediction using Random Forest},
  journal   = {Engineering And Technology Journal (ETJ)},
  volume    = {10},
  number    = {5},
  pages     = {5159--5164},
  year      = {2025},
  doi       = {10.47191/etj/v10i05.44},
  url       = {https://everant.org/index.php/etj/article/view/1939}
}

@article{Marjani2023FirePred,
  author    = {Mohammad Marjani and Seyed Ali Ahmadi and Masoud Mahdianpari},
  title     = {FirePred: A hybrid multi-temporal convolutional neural network model for wildfire spread prediction},
  journal   = {Ecological Informatics},
  volume    = {78},
  pages     = {102282},
  year      = {2023},
  doi       = {10.1016/j.ecoinf.2023.102282},
  url       = {https://www.sciencedirect.com/science/article/pii/S1574954123003114}
}

@article{Pérez-Porras2021LargeWildfires,
  author    = {Fernando-Juan Pérez-Porras and Paula Triviño-Tarradas and Carmen Cima-Rodríguez and Jose-Emilio Meroño-de-Larriva and Alfonso García-Ferrer and Francisco-Javier Mesas-Carrascosa},
  title     = {Machine learning methods and synthetic data generation to predict large wildfires},
  journal   = {Sensors},
  volume    = {21},
  number    = {11},
  pages     = {3694},
  year      = {2021},
  doi       = {10.3390/s21113694},
  url       = {https://www.ncbi.nlm.nih.gov/pmc/articles/PMC8198242/}
}

@misc{Angelov2021WildfireSizeBinary,
  author       = {Dimitar Milenov Angelov},
  title        = {Machine Learning for Binary Classification of Wildfire Size},
  howpublished = {Bachelor thesis, Tilburg University},
  year         = {2021},
  note         = {Available online: http://arno.uvt.nl/show.cgi?fid=156918},
  url          = {http://arno.uvt.nl/show.cgi?fid=156918}
}

@misc{caron2025localized,
  title={Localized Forest Fire Risk Prediction: A Department-Aware Approach for Operational Decision Support},
  author={Caron, Nicolas and Guyeux, Christophe and Noura, Hassan and Aynes, Benjamin},
  year={2025},
  eprint={2506.04254},
  archivePrefix={arXiv},
  primaryClass={cs.LG},
  url={https://arxiv.org/abs/2506.04254}
}

@article{Syphard2024_geography_fire_patterns,
  title        = {The importance of geography in forecasting future fire patterns under climate change},
  author       = {Syphard, Alexandra D. and Velazco, Santiago Jos{\'e} El{\'\i}as and Rose, Miranda Brooke and Franklin, Janet and Regan, Helen M.},
  journal      = {Proceedings of the National Academy of Sciences of the United States of America},
  volume       = {121},
  number       = {32},
  pages        = {e2310076121},
  year         = {2024},
  doi          = {10.1073/pnas.2310076121},
  pmcid        = {PMC11317612},
  url          = {https://doi.org/10.1073/pnas.2310076121}
}

@article{AlBashiti2022_machineLearningWildfire,
  title        = {Machine Learning for Wildfire Classification: Exploring blackbox, eXplainable, symbolic, and SMOTE methods},
  author       = {Al-Bashiti, Mohammad Khaled and Naser, M. Z.},
  journal      = {Natural Hazards Research},
  volume       = {2},
  number       = {3},
  pages        = {154--165},
  year         = {2022},
  doi          = {10.1016/j.nhres.2022.08.001},
  url          = {https://doi.org/10.1016/j.nhres.2022.08.001}
}

@article{Wang2020_wildfire_burned_area,
  title        = {Quantifying the effects of environmental factors on wildfire burned area in the south central US using integrated machine learning techniques},
  author       = {Wang, Sally S.-C. and Wang, Yuxuan},
  journal      = {Atmospheric Chemistry and Physics},
  volume       = {20},
  pages        = {11065--11087},
  year         = {2020},
  doi          = {10.5194/acp-20-11065-2020},
  url          = {https://acp.copernicus.org/articles/20/11065/2020/}
}

@article{Coffield2019_fire_size_prediction,
  title        = {Machine learning to predict final fire size at the time of ignition},
  author       = {Coffield, Shane R. and Graff, Casey A. and Chen, Yang and Smyth, Padhraic and Foufoula-Georgiou, Efi and Randerson, James T.},
  journal      = {International Journal of Wildland Fire},
  volume       = {28},
  number       = {11},
  pages        = {861--873},
  year         = {2019},
  doi          = {10.1071/WF19023},
  pmcid        = {PMC8152111},
  url          = {https://pmc.ncbi.nlm.nih.gov/articles/PMC8152111/}
}

@article{Li2023_AttentionFire,
  title        = {AttentionFire\_v1.0: interpretable machine learning fire model for burned-area predictions over tropics},
  author       = {Li, Fa and Zhu, Qing and Riley, William J. and Zhao, Lei and Xu, Li and Yuan, Kunxiaojia and Chen, Min and Wu, Huayi and Gui, Zhipeng and Gong, Jianya and Randerson, James T.},
  journal      = {Geoscientific Model Development},
  volume       = {16},
  pages        = {869--884},
  year         = {2023},
  doi          = {10.5194/gmd-16-869-2023},
  url          = {https://gmd.copernicus.org/articles/16/869/2023/}
}

@article{Balch2017_humanStartedWildfires,
  title        = {Human-started wildfires expand the fire niche across the United States},
  author       = {Balch, Jennifer K. and Bradley, Bethany A. and Abatzoglou, John T. and Nagy, R. Chelsea and Fusco, Emily J. and Mahood, Adam L.},
  journal      = {Proceedings of the National Academy of Sciences of the United States of America},
  volume       = {114},
  number       = {11},
  pages        = {2946--2951},
  year         = {2017},
  doi          = {10.1073/pnas.1617394114},
  url          = {https://www.pnas.org/doi/10.1073/pnas.1617394114}
}

@article{Fusco2022_detectionBiasesWildfire,
  author = {Rodriguez-Cubillo, Dario and Pilon, Natashi A. L. and Durigan, Giselda},
    title = {Tree height is more important than bark thickness, leaf habit or habitat preference to survive fire in the cerrado of south-east Brazil},
    journal = {International Journal of Wildland Fire},
    volume = {30},
    number = {11},
    pages = {899-910},
    year = {2021},
    month = {09},
    issn = {1049-8001},
    doi = {10.1071/WF21091},
    url = {https://doi.org/10.1071/WF21091},
    eprint = {https://connectsci.au/wf/article-pdf/30/11/899/737449/wf21091.pdf},
}

@inproceedings{chen2016xgboost,
  title={XGBoost: A Scalable Tree Boosting System},
  author={Chen, Tianqi and Guestrin, Carlos},
  booktitle={Proceedings of the 22nd ACM SIGKDD International Conference on Knowledge Discovery and Data Mining},
  series={KDD '16},
  year={2016},
  pages={785--794},
  publisher={ACM},
  address={New York, NY, USA},
  doi={10.1145/2939672.2939785},
  url={https://dl.acm.org/doi/10.1145/2939672.2939785}
}

@article{hochreiter1997long,
  title={Long Short-Term Memory},
  author={Hochreiter, Sepp and Schmidhuber, Jürgen},
  journal={Neural Computation},
  volume={9},
  number={8},
  pages={1735--1780},
  year={1997},
  publisher={MIT Press},
  doi={10.1162/neco.1997.9.8.1735},
  url={https://doi.org/10.1162/neco.1997.9.8.1735}
}

@article{Paulus2022_DataCognitiveBiasCrisisInfo,
  title        = {On the Interplay of Data and Cognitive Bias in Crisis Information Management: An Exploratory Study on Epidemic Response},
  author       = {Paulus, David and Fathi, Ramian and Fiedrich, Frank and Van de Walle, Bartel and Comes, Tina},
  journal      = {Information Systems Frontiers},
  volume       = {26},
  number       = {2},
  pages        = {391--415},
  year         = {2022},
  doi          = {10.1007/s10796-022-10241-0},
  url          = {https://doi.org/10.1007/s10796-022-10241-0}
}

@article{FriedGilless1989_firelineProduction,
  title        = {Expert opinion estimation of fireline production rates},
  author       = {Fried, Jeremy S. and Gilless, J. Keith},
  journal      = {Forest Science},
  volume       = {35},
  number       = {3},
  pages        = {870--877},
  year         = {1989},
  doi          = {10.1093/forestscience/35.3.870},
  url          = {https://academic.oup.com/forestscience/article/35/3/870/4642033}
}

@article{HirschCoreyMartell1998_fireCrewEffectiveness,
  title        = {Using Expert Judgment to Model Initial Attack Fire Crew Effectiveness},
  author       = {Hirsch, Kelvin G. and Corey, Paul N. and Martell, David L.},
  journal      = {Forest Science},
  volume       = {44},
  number       = {4},
  pages        = {539--549},
  year         = {1998},
  doi          = {10.1093/forestscience/44.4.539},
  url          = {https://academic.oup.com/forestscience/article/44/4/539/4627517}
}

@article{Zigner2020_FARSITE_SantaBarbara,
  title        = {Evaluating the Ability of FARSITE to Simulate Wildfires Influenced by Extreme, Downslope Winds in Santa Barbara, California},
  author       = {Zigner, Katelyn and Carvalho, Leila M. V. and Peterson, Seth and Fujioka, Francis and Duine, Gert‑Jan and Jones, Charles and Roberts, Dar and Moritz, Max},
  journal      = {Fire},
  volume       = {3},
  number       = {3},
  pages        = {29},
  year         = {2020},
  doi          = {10.3390/fire3030029},
  url          = {https://doi.org/10.3390/fire3030029}
}

@inproceedings{Finney2006_FlamMapOverview,
  author    = {Finney, Mark A.},
  title     = {An Overview of FlamMap Fire Modeling Capabilities},
  booktitle = {Fuels Management -- How to Measure Success: Conference Proceedings},
  editors   = {Andrews, Patricia L. and Butler, Bret W.},
  series    = {RMRS-P-41},
  year      = {2006},
  pages     = {213--220},
  address   = {Portland, OR / Fort Collins, CO},
  publisher = {U.S. Department of Agriculture, Forest Service, Rocky Mountain Research Station},
  url       = {https://research.fs.usda.gov/treesearch/25948}
}

@INPROCEEDINGS{Pham2022CaliforniaWildfireML,
  author={Pham, Kaylee and Ward, David and Rubio, Saulo and Shin, David and Zlotikman, Lior and Ramirez, Sergio and Poplawski, Tyler and Jiang, Xunfei},
  booktitle={2022 21st IEEE International Conference on Machine Learning and Applications (ICMLA)}, 
  title={California Wildfire Prediction using Machine Learning}, 
  year={2022},
  volume={},
  number={},
  pages={525-530},
  doi={10.1109/ICMLA55696.2022.00086}}

@article{Peterson2011LongTermRegimes,
  author  = {Peterson, S. H. and Moritz, M. A. and Morais, M. E. and Dennison, P. E. and Carlson, J. M.},
  year    = {2011},
  title   = {Modelling Long-Term Fire Regimes of Southern California Shrublands},
  journal = {International Journal of Wildland Fire},
  volume  = {20},
  pages   = {1--16},
  doi     = {10.1071/WF09102},
  url     = {https://doi.org/10.1071/WF09102},
}

@article{Preisler2004WildfireRisk,
  title        = {Probability based models for estimation of wildfire risk},
  author       = {Preisler, Haiganoush K. and Brillinger, David R. and Burgan, Robert E. and Benoit, John W.},
  journal      = {International Journal of Wildland Fire},
  volume       = {13},
  number       = {2},
  pages        = {133--142},
  year         = {2004},
  doi          = {10.1071/WF02061},
  url          = {https://doi.org/10.1071/WF02061}
}

@misc{CALFIRE_HistoricalPerimeters,
  author       = {{California Department of Forestry and Fire Protection (CAL FIRE) — FRAP}},
  title        = {California Historical Fire Perimeters},
  howpublished = {Data set, Fire and Resource Assessment Program (FRAP)},
  year         = {2025},
  url          = {https://data.ca.gov/dataset/california-historical-fire-perimeters}
}

@inproceedings{Pham2022_CAWildfirePrediction,
  author    = {Pham, Kaylee and Ward, David M. and Rubio, Saulo and Shin, David and Zlotikman, Lior and Ramirez, Sergio and Poplawski, Tyler and others},
  title     = {California Wildfire Prediction using Machine Learning},
  booktitle = {2022 21st IEEE International Conference on Machine Learning and Applications (ICMLA)},
  year      = {2022},
  pages     = {525--530},
  url       = {https://www.csun.edu/~xjiang/SeniorDesign/sd2021/BeatTheHeat/assets/Wildfire%20Predictions%20in%20California.pdf},
  note      = {Accessed: 2025-11-30}
}

@article{Tzimas2024ForestFireDuration,
  title        = {Predicting the Duration of Forest Fires Using Machine Learning Methods},
  author       = {Kopitsa, Constantina and Tsoulos, Ioannis G. and Charilogis, Vasileios and Stavrakoudis, Athanassios},
  journal      = {Future Internet},
  volume       = {16},
  number       = {11},
  pages        = {396},
  year         = {2024},
  doi          = {10.3390/fi16110396},
  url          = {https://www.mdpi.com/1999-5903/16/11/396}
}

@misc{CALFIRE_PerimetersAll,
  author       = {{CAL FIRE}},
  title        = {California Fire Perimeters (all)},
  howpublished = {Data set},
  url          = {https://data.ca.gov/dataset/california-fire-perimeters-all},
}

@misc{FRAP_DataDictionary,
  author       = {{CAL FIRE Fire and Resource Assessment Program}},
  title        = {FRAP Historic Fire Perimeters Data Dictionary},
  howpublished = {Data dictionary},
  url          = {https://34c031f8-c9fd-4018-8c5a-4159cdff6b0d-cdn-endpoint.azureedge.net/-/media/calfire-website/what-we-do/fire-resource-assessment-program---frap/historic-fire-perimeters-data-dictionary.pdf?rev=4c13140f376d4a7782722338cf5b13ef&hash=E6392C1F2A65CA13AE2B4DF4FBE3AAB4},
}

@misc{CaliforniaOpenData_Shapefile,
  author       = {{California Open Data Portal}},
  title        = {California Fire Perimeters (all) Shapefile},
  howpublished = {Shapefile data set},
  url          = {https://data.ca.gov/dataset/california-fire-perimeters-all/resource/654e8eca-4b7e-4a55-9165-577f586e0042},
}

@article{Li2021SpatialTemporalCA,
  author  = {Li, S. and Banerjee, T.},
  year    = {2021},
  title   = {Spatial and Temporal Pattern of Wildfires in California from 2000 to 2019},
  journal = {Scientific Reports},
  volume  = {11},
  eid     = {8779},
  doi     = {10.1038/s41598-021-88131-9},
  url     = {https://doi.org/10.1038/s41598-021-88131-9},
}

\appendix

\section{AI Use}

Artificial intelligence tools (such as Overleaf AI assistance) were used only for grammar and spelling correction. All research, writing, analysis, and interpretation reflect the author’s work.

\end{document}